\documentclass[runningheads]{llncs}
\usepackage[T1]{fontenc}

\usepackage{graphicx} 
\usepackage{hyperref}

\title{Learning a Distance for the \guillaume{Clustering}  of Patients with Amyotrophic Lateral Sclerosis}
\author{Guillaume Tejedor\inst{1}\orcidID{0009-0006-9537-4735}   \and
        Veronika Peralta\inst{1}\orcidID{0000-0002-9236-9088}    \and
        Nicolas Labroche\inst{1}\orcidID{0000-0002-2794-2124}    \and
        Patrick Marcel\inst{2}\orcidID{0000-0003-3171-1174}      \and
        Hélène Blasco\inst{3,4}\orcidID{0000-0001-6107-0035}     \and
        Hugo Alarcan\inst{3, 4}\orcidID{0000-0001-8707-1722}}

\institute{
    Laboratoire LIFAT, Université de Tours, firstname.name@univ-tours.fr \and
    Laboratoire LIFO, Université d'Orléans, patrick.marcel@univ-orleans.fr \and
    Service de Biochimie et Biologie Moléculaire CHRU de Tours, firstname.name@univ-tours.fr \and
    UMR U1253 iBrain, Université de Tours, Inserm\\firstname.name@univ-tours.fr
}

\usepackage{graphicx}
\usepackage{xcolor}

\usepackage[T1]{fontenc}
\usepackage[utf8]{inputenc}
\usepackage{ifxetex}
\usepackage{url}
\usepackage{lmodern}
\usepackage{bm}
\usepackage{array} 
\usepackage{longtable}
\usepackage{multirow}
\usepackage{floatrow}
\usepackage{float}
\floatstyle{plaintop}
\restylefloat{table}

\usepackage{url}

\newcommand{\vero}[1]{\textcolor{black}{#1}}
\newcommand{\guillaume}[1]{\textcolor{black}{#1}}

\date{April 2025}

\begin{document}
    \maketitle
    \setcounter{footnote}{0}

    \begin{abstract}
        Amyotrophic lateral sclerosis (ALS) is a severe disease with a typical survival of 3–5 years after symptom onset. Current treatments offer only limited life extension, and the variability in patient responses highlights the need for personalized care. However, research is hindered by small, heterogeneous cohorts, sparse longitudinal data, and the lack of a clear definition for clinically meaningful patient \guillaume{clusters}. 
        Existing \guillaume{clustering} 
        methods remain limited in both scope and number.
        \guillaume{To address this, we propose a clustering approach that groups sequences using a disease progression declarative score.}
        Our approach integrates medical expertise through multiple descriptive variables, investigating several distance measures combining such variables, both by reusing off-the-shelf distances and employing a weak-supervised learning method. We pair these distances with clustering methods and benchmark them against state-of-the-art techniques. The evaluation of our approach on a dataset of 353 ALS patients from the University Hospital of Tours, shows that our method outperforms state-of-the-art methods in survival analysis while achieving comparable silhouette scores. In addition, the learned distances enhance the relevance and interpretability of results for medical experts.
    \end{abstract}
    
    \section{Introduction}
    \label{sec:intro}
    
    Amyotrophic lateral sclerosis (ALS) is a fatal neurodegenerative disease of the central nervous system that is difficult to diagnose, especially in its early stages \cite{feldman2022amyotrophic}. 

    It primarily affects motor neurons in the brain and spinal cord, which are responsible for controlling voluntary muscle movements.
    ALS research faces significant challenges due to factors such as small, heterogeneous patient cohorts, limited longitudinal data, and the lack of clear definitions for clinically meaningful patient \guillaume{clusters} 
    (phenotypes) and ground truth. These issues, compounded by data sparsity, make it difficult to develop reliable computational models for diagnosis and prognosis. Datasets are often constrained by costly, invasive procedures and strict data-sharing regulations, requiring substantial resources to maintain high-quality data. As a result, the available samples tend to be scarce and fragmented, which strains machine learning models
    \cite{grollemund2019machine}
    and weakens their generalization, further complicating efforts to identify distinct patient \guillaume{clusters},
    develop personalized care options and improve early-stage treatment.

    \begin{figure}[t] 
        \centering 
        \includegraphics[height=0.2\textheight]{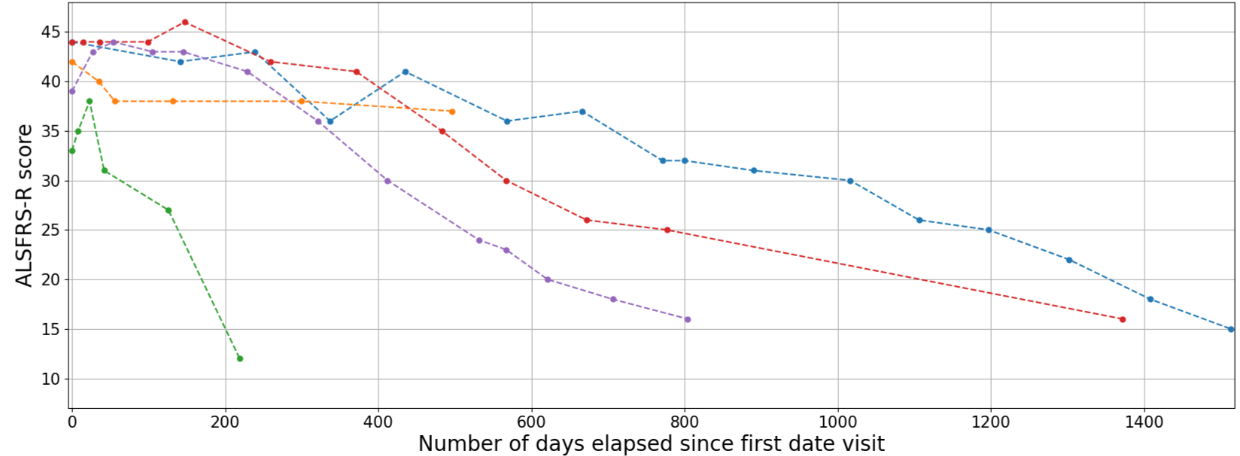} 
        \caption{Example of five ALSFRS-R sequences. 
        }
        \label{fig:sequence_examples}
    \end{figure}
    
    To identify \guillaume{clusters}, most existing studies use longitudinal clinical data to represent disease progression 
    \cite{tavazzi2023artificial}. The “ALS Functional Rating Scale-Revised” (ALSFRS-R) 
    remains the standard reference for measuring disease progression \cite{feldman2022amyotrophic}. This scale evaluates a patient's physical function using 12 questions, each representing a distinct physical function  (e.g., Speech, Salivation) and scored from 0 to 4. The total score ranges from 0 (total physical function loss) to 48 (no functional impairment). It is a revised version of the original ALSFRS \guillaume{score}, which had only 10 questions. Fig. \ref{fig:sequence_examples} shows examples of temporal sequences depicting ALSFRS-R progression for 5 patients.
    
    In the literature, some studies assign \guillaume{clusters} 
    using threshold rules applied to the ALSFRS-R score at 
    specific time points. This approach is limited because it relies on a single criterion and is highly sensitive to small score variations around the thresholds, which may not reflect meaningful clinical differences. Moreover, these methods do not consider multiple criteria or the temporal progression of the disease, which could improve the robustness of 
    \guillaume{cluster} assignment \cite{halbersberg2019temporal}.
    
    Without assuming prior identification of specific \guillaume{clusters}, we propose an approach for patient \guillaume{clustering} 
    that learns a distance measure using a weak-supervised method. In addition, we propose to test different off-the-shelf distances. Theses distances are then paired with clustering methods. Drawing inspiration from clustering of Business Intelligence sequences
    \cite{drushku2019interest}, we develop distances for patients’ ALSFRS-R sequences using descriptive variables extracted from these temporal data. The goal is to use these measures in a way that captures patterns of disease progression, and uncover distinct patient \guillaume{clusters}, each exhibiting unique forms and rates of progression.
    
    Precisely, the contributions presented in this article include:
    \begin{enumerate}
        \item The identification of a set of descriptive variables for characterizing disease progression within ALSFRS-R sequences.
        \item The development of distances based on the extracted variables.
        \item An approach to automatically discover patient \guillaume{clusters}
        based on theses distances 
        and off-the-shelf clustering methods.
    \end{enumerate}
    
    The remainder of the paper is structured as follows: Section \ref{sec:sota} presents the current state of research on ALS patient
    \guillaume{clustering} using longitudinal clinical data. Section \ref{sec:approach} describes our approach. Section \ref{sec:experiments} details experiments and results, comparing our approach to state-of-the-art methods. Finally, Section \ref{sec:conclusion} provides our conclusions and perspectives.
    
    \section{Related work}
    \label{sec:sota}

    In this section, we review the proposals  
    for 
    ALS patient \guillaume{clustering} using the ALSFRS-R score and the available datasets.
    
    \subsection{
    \guillaume{Clustering} based on progression thresholds}
    
        One common approach to 
        \guillaume{cluster patients} is to observe ALSFRS-R score at a specific time point 
        and use threshold-based rules to define 
        \guillaume{clusters}. For example, Gomeni et al. calculate the percentage of change from first ALSFRS-R score after six months \cite{gomeni2014amyotrophic}.
        They define two 
        \guillaume{clusters}: one representing patients with slow disease progression (i.e., decline 
        < 18.6\%) and fast disease progression (i.e., decline >= 18.6\%). The value of 18.6\% is determined by computing the \guillaume{$90^{th}$} 
        percentile of the distribution of ALSFRS-R \guillaume{score} at six months.
        
        Grollemund et al. \cite{grollemund2021manifold} propose a similar approach by defining four 
        \guillaume{clusters} directly based on the ALSFRS score one year after the start of follow-up: score $\leq$ 10, 10 < score $\leq$ 20, 20 < score $\leq$ 30, and score > 30. 
        
        Meyer et al. \cite{meyer2025phosphorylated}
        use
        the D50 model, which
        outputs the number of months from the start of follow-up where
        the patient's ALSFRS-R score drops to 24, indicating a halving of functional ability.
        They define three 
        \guillaume{clusters} corresponding to different levels of disease severity: severe (D50 $<$ 20), intermediate (20 $\leq$ D50 $<$ 40), and mild (D50 $\geq$ 40) \cite{meyer2025phosphorylated}.
        However, using approaches that reduce an entire sequence to a single value limits the ability to identify diverse 
        \guillaume{clusters}. Moreover, the heterogeneity of ALS implies that a single criterion may not fully characterize the disease.
        
    \subsection{Sequence comparison measures}
    
        To our knowledge, only Halbersberg et al. propose a comparison measure for evaluating pairs of ALSFRS sequences \cite{halbersberg2019temporal}. Their approach relies on multivariate ALSFRS sequences, where each time point includes 10 dimensions corresponding to ALSFRS subscores
        (e.g., Speech, Salivation), individually rated on a scale from 0 to 4.
        To compare pairs of sequences, they use the Dynamic Time Warping (DTW) dissimilarity \cite{sakoe1978dynamic}. 
        They propose the independent warping approach $DTW_I$, which extends DTW to multi-dimensional sequences. Specifically,
        the dissimilarity between two multi-dimensional
        sequences is calculated as the sum of the DTW dissimilarities computed independently for each dimension \cite{shokoohi2017generalizing}. Formally, given two patient sequences $S$ and $Q$, $DTW_I$ is computed as follows:
        
        $$DTW_{I}(S,Q)=\sum_{i=1}^{10} DTW(S_{i}, Q_{i})$$
        \noindent
        where $S_i$ and $Q_i$ denote the \guillaume{$i^{th}$} subscore of sequences $S$ and $Q$.

        After computing $DTW_{I}$ dissimilarity for each pair of sequences, a dissimilarity matrix is constructed and hierarchical clustering is applied. The number of clusters is set between 4 and 6 and a minimal number of 80 patients in each cluster is sought. The resulting clusters are then used for a classification task that predicts the patient's ALSFRS score at the next visit. 
        
        While Halbersberg et al. present promising classification results, they do not evaluate the quality of the clusters using clustering metrics. Furthermore, they do not justify why DTW would be more appropriate than other comparison measures 
        \guillaume{to cluster} patients based on ALSFRS sequences. Additionally, DTW was originally developed to align two time series of unequal lengths, under the assumption that the sequences are synchronous with a constant time interval between visits. \cite{sakoe1978dynamic}. However, in most ALS datasets, patient visits occur asynchronously at irregular intervals, making DTW not suitable in such contexts.
        
    \subsection{Available datasets}
    
        Few ALS 
        datasets are publicly available online, the most widely used being PRO-ACT, which includes over 10,000 patients and offers a wide variety of clinical and biological parameters. Other datasets from hospitals in Paris (France) and Jena (Germany) are also available but include significantly fewer patients \cite{tavazzi2023artificial}.
        
        However, PRO-ACT has several limitations. The patients come from 23 different clinical trials, which leads to inconsistencies in follow-up frequency and unit conventions. Additionally, the dataset only includes information starting from the point of diagnosis, with no data about first follow-up visits.
        Finally, there is no way to contact the clinicians who followed the patients, which limits the contextual understanding of the data. For these reasons, we consider to work with a dataset from a hospital where we can directly collaborate with physicians who possess expert knowledge of the data and its clinical context.
        
   \subsection{Summary}
   
         All contributions are summarized in Table \ref{tab:contributions_info}. The table presents the \guillaume{clustering} 
         approaches,
         number of clusters, used dataset(s) and number of patients per cluster. The first column contains acronyms identifying
         each paper: GOM for Gomeni et al., GRO for Grollemund et al., MEY for Meyer et al., and HAL for Halbersberg et al.. They are also used in
         experiments in Section \ref{sec:experiments}.
  
        \begin{table}[t]
            \caption{Overview of the datasets used, number of patients after preprocessing, the 
            \guillaume{clustering} methods applied, and the number of clusters identified by state-of-the-art techniques.}
            \centering
            \resizebox{1\textwidth}{!}{
            \begin{tabular}{|c|>{\centering\arraybackslash}p{4cm}|>{\centering\arraybackslash}p{2.5cm}|>{\centering\arraybackslash}p{3cm}|>{\centering\arraybackslash}p{2cm}|}
            \hline
            \textbf{Id} & \textbf{Approach} & \textbf{Number of clusters} & \textbf{Dataset(s)} & \textbf{Number of patients} \\ \hline
            
            GOM & Threshold: \% change from baseline at six months \cite{gomeni2014amyotrophic} & 2 & PRO-ACT & 338 \\ \hline
            
            \multirow{4}{*}{GRO} & \multirow{4}{*}{\parbox{4cm}{\centering Threshold: ALSFRS score after one year \cite{grollemund2021manifold}}} & \multirow{4}{*}{4} & Trophos & 357 \\ \cline{4-5}
            & & & Exonhit & 227 \\ \cline{4-5}
            & & & PRO-ACT & 2841 \\ \cline{4-5}
            & & & Paris tertiary referral centre & 331 \\ \hline
            
            MEY & Threshold: D50 value \cite{meyer2025phosphorylated} & 3 & Jena University Hospital & 108 \\ \hline
            
            HAL & DTW independent \cite{halbersberg2019temporal} & 4 to 6 & PRO-ACT & 3171 \\ \hline
            \end{tabular}
            }
            \label{tab:contributions_info}
        \end{table}

        For the PRO-ACT dataset, the number of patients may vary depending on the data preparation steps outlined in the respective studies.
        Finally, we remark that none of the papers use internal metrics to evaluate how well-separated the clusters are. This omission may be limiting, as metrics like silhouette score can offer valuable insights about the intrinsic quality of the clustering and help determine whether the discovered structure is genuinely meaningful.
  
    \section{
\guillaume{Clustering} approach}
    \label{sec:approach}
    We propose to represent the disease progression of a patient
    as a temporal sequence <$s_{t_0}$... $s_{t_{n}}$>, where $s_{t_i}$ denotes the ALSFRS-R score measured within a follow-up
    visit at relative time $t_i$ (measured in days elapsed from the first visit), with $t_0 = 0$ and $t_{n}$ corresponding to the last visit time. Due to the decreasing score and increasing time, $s_{t_i} \geq s_{t_{i+1}}$ except in rare cases.

    \begin{figure}[ht]
        \centering
        \includegraphics[height=0.15\textheight]
        {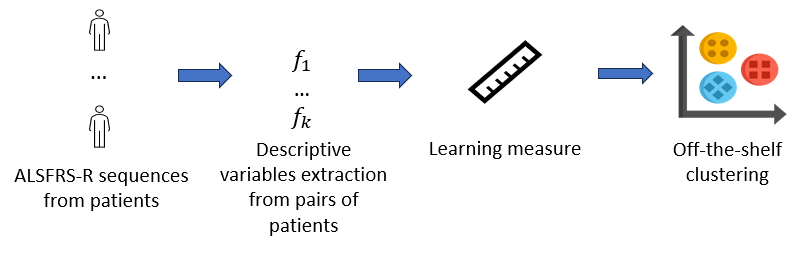}
        \caption{Framework describing the different steps of the clustering process. 
        }
        \label{fig:stratification_approach}
    \end{figure}

    Figure \ref{fig:stratification_approach} presents our general framework. 
    We take as input the ALSFRS-R sequences of a set of patients. For each pair of sequences,
    we extract descriptive variables that characterize their differences (detailed in Subsection \ref{sec:extracted_features}).  
    These variables are then used to capture the distance between
    each pair. We investigate several distances, from classic ones to 
    more advanced methods based on weak-supervised learning (explained
    in Section \ref{sec:learning_measures}). The proposed
    measures are paired with off-the-shelf 
    clustering methods \guillaume{(i.e., k-means, k-medoids, hierarchical clustering)}, 
    enabling the identification of patient 
    \guillaume{clusters}. 

    \subsection{Extraction of
    descriptive variables}
    \label{sec:extracted_features}
        \label{sec:features_extraction}
        
\begin{figure}[t]
        \centering
        \includegraphics[height=0.20\textheight]
        {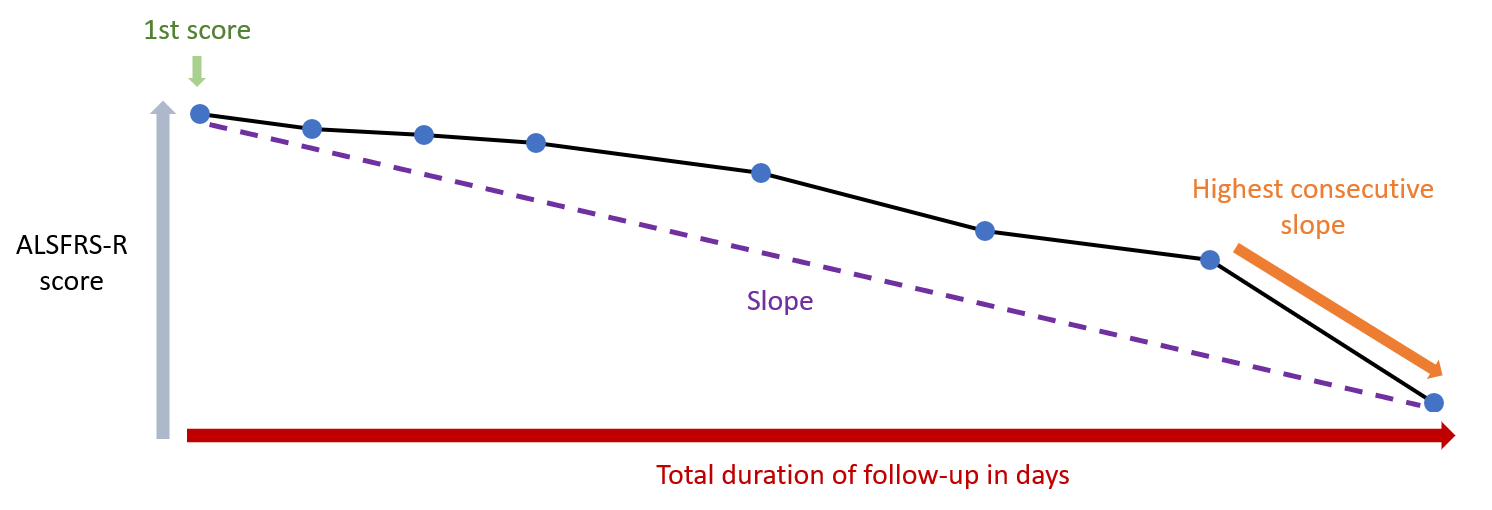}
        \caption{Extracted information from the ALSFRS-R sequences of each patient.}
        \label{fig:features_whole_sequence}
    \end{figure}

\begin{figure}[t]
        \centering
        \includegraphics[height=0.30\textheight]{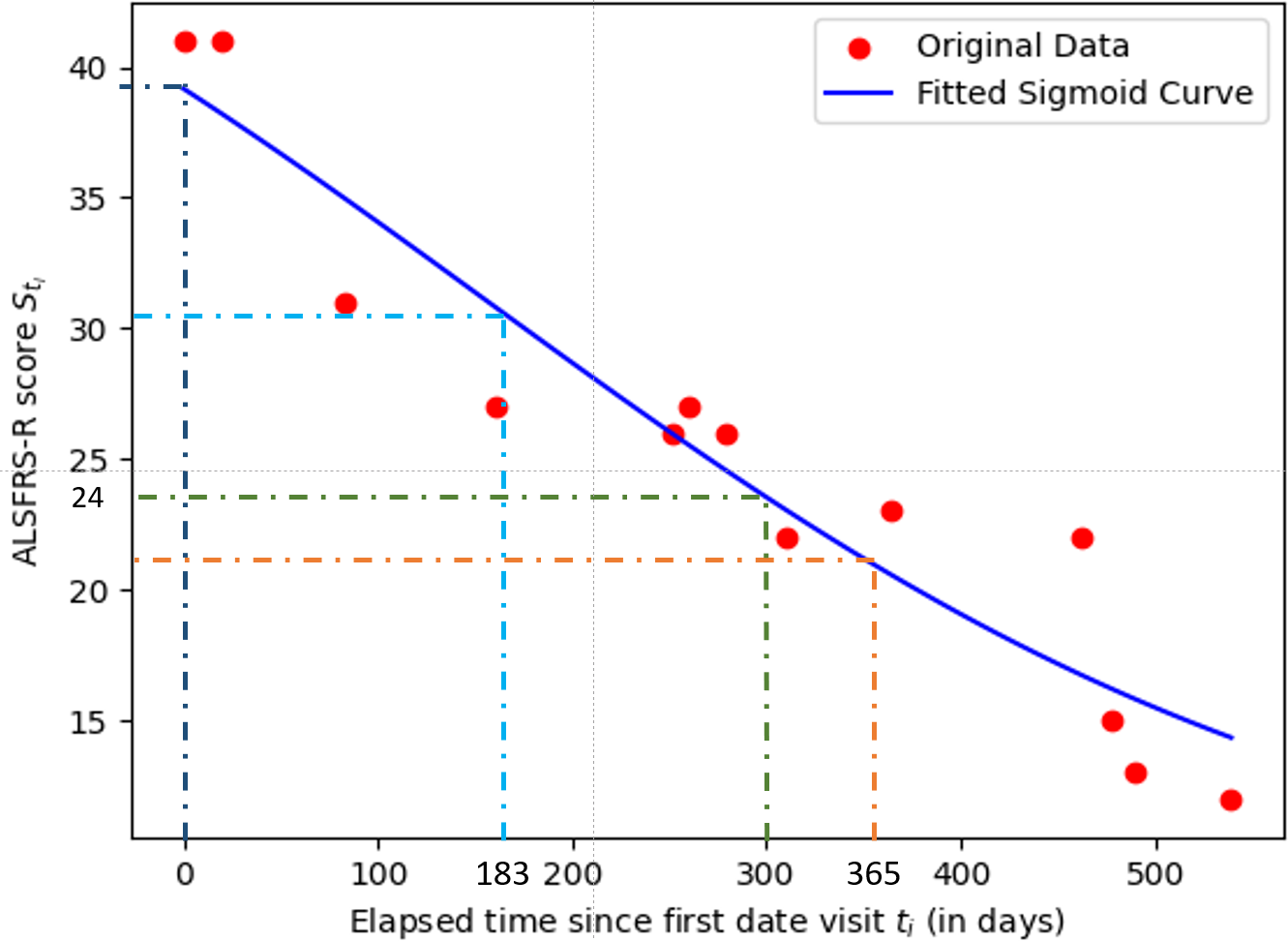}
        \caption{Extracted information from a fitted sigmoid on an ALSFRS-R sequence. $S_{365}$ is the one-year score, $S_0$ and $S_{183}$ are the initial and six-month scores used to compute the percentage of decrease, and $S_{300}$ marks when the score reaches 24 (D50) \cite{meyer2025phosphorylated}.}
        \label{fig:fitted_sigmoid_example}
    \end{figure}

    In order to define a set of
    descriptive variables for each pair of patients, we 
    extract relevant features
    from ALSFRS-R
    sequences, as illustrated in Fig. \ref{fig:features_whole_sequence} and 
    \ref{fig:fitted_sigmoid_example}.

        \paragraph{\textbf{Sequence-based feature extraction}}: Based on feedback from physicians, we extract the following features from 
        each ALSFRS-R sequence:
        the total follow-up duration; the score at the first visit; the overall slope of the sequence, which is the difference in ALSFRS-R scores between the first and last visits divided by the total follow-up duration, i.e., $\frac {s_{t_{n}} - s_{t_{0}}} {t_{n}}$; and the stiffest slope between two consecutive visits, i.e., $\min_{0 \leq i \leq {n}-1} \frac {s_{t_{i+1}} - s_{t_i}} {t_{i+1} - t_i}$ (see Fig. \ref{fig:features_whole_sequence}).

        \paragraph{\textbf{Sigmoid-based feature extraction}}: We extract additional features
        inspired by related works discussed in Section~\ref{sec:sota}. Specifically, we consider the ALSFRS-R score after one year of follow-up, i.e., $s_{t_i}$ at 
        $t_i=365$; the percentage of decline
        after six months, i.e., $\frac{s_{t_0}-s_{t_i}}{s_{t_0}}$ at 
        $t_i=183$; and the D50 value (in days), i.e., $t_i$ where $s_{t_i}=24$
        (see Fig. \ref{fig:fitted_sigmoid_example}).

        Remark that
        since the sequences are not synchronized, there is not necessarily a visit at (or close to) time 183 and 365, nor a visit observing  
        an ALSFRS-R score 
        equal to 24.
        To address this, we reuse the state-of-the-art strategy of fitting
        a sigmoid function \cite{meyer2025phosphorylated} to estimate these values, with the sigmoid parameters
        being learned over multiple runs. The formulation is as follows:

        $$ sigmoid(x, b, m, a, c) = \frac{b}{1 + e^{m(x - a)}} + c $$
        
        \noindent
        where $x$ represents the number of days elapsed since the start of the follow-up, and $b$, $m$, $a$, and $c$ are the parameters to be learned. This function returns an estimated ALSFRS-R score at time $x$.
    
        In a preliminary study, we evaluated several sigmoid parameter configurations, ranging from 1 to 5 parameters, and selected this one because it yields the lowest Root Mean Squared Error (RMSE).
        
        From all these features, we
        compute a set of 
        descriptive variables describing the difference between a pair of ALSFRS-R sequences,
        namely:
        
        \begin{enumerate}
            \item DURATION\_DIFF: The absolute difference between the total follow-up durations of the two
            sequences.
            \item FIRST\_SCORE\_DIFF: The absolute difference between the first ALSFRS-R scores of the two sequences.
            \item SLOPE\_DIFF: The absolute difference between the overall slopes of the two
            sequences.
            \item HIGHEST\_CONSECUTIVE\_SLOPE\_DIFF: The absolute difference between the stiffest
            consecutive slopes of the two sequences.
            \item ALS\_SCORE\_M12\_DIFF: The absolute difference between the ALSFRS-R scores after one year of follow-up of the two sequences.
            \item PC\_CHANGE\_M6\_DIFF: The absolute difference between percentage of change after six months of follow-up of the two 
            sequences.
            \item D50\_DIFF: The absolute difference between the D50 values of the two 
            sequences.
        \end{enumerate}

        \guillaume{Each descriptive variable is normalized between 0 and 1 using the Min-Max Scaler. This method was chosen to ensure all variables share the same scale and magnitude within the [0, 1] range while preserving the data’s shape, allowing differences to be interpreted as distances.}
        
        Spearman correlation is used to filter out correlated \guillaume{descriptive variables}\vero{, when} 
        \guillaume{ 
        correlation coefficient is greater than 
        0.7. 
        } 

\subsection{Proposed distances}
    \label{sec:learning_measures}
    
    In this subsection, we investigate several distances
    to compare 
    two patients $P_X$ and $P_Y$,
    and be paired with  
    clustering methods.
    We consider some standard distances commonly used with \guillaume{such} 
    clustering methods.
    We also present a weak-supervised approach that combines a generative model with a classifier. 
    
    In the following, we 
    consider two vectors: \( X = (x_1, \ldots, x_k) \) and \( Y = (y_1, \ldots, y_k) \), 
    each $x_i$ (resp. $y_i$) corresponding to the $i^{th}$ feature computed for the ALSFRS-R sequence of patient $P_X$ (resp. $P_Y$), with $k=7$. Our descriptive variable is represented as $|x_{i}-y_{i}|$ where $1<=i<=k$.
    
    \paragraph{\textbf{Off-the-shelf distances:}} We propose two classic distances: Minkowski and Cosine distances.
    
    \begin{itemize}

    \item \textbf{Minkowski distance}
    of order p (with p $\geq$ 1) forms a set of metrics within a vector space. It is defined such that:
    $$D_p(X, Y) = \left( \sum_{i=1}^k |x_i - y_i|^p \right)^{\frac{1}{p}}$$
    We compute the Minkowski distance of order \( p = 1 \) and \( p = 2 \), 
    corresponding to Manhattan (MAN) and Euclidean (EUC) distances, respectively.
    
    \item \textbf{Cosinus distance}
    (COS) measures one minus the angle between two vectors, the latter computed as
    the dot product of the vectors divided by the product of their magnitudes. It is defined such that:
    $$COS(X, Y) = 1 - \frac{X \cdot Y}{\|X\| \, \|Y\|}$$
        
    \end{itemize}

    \paragraph{\textbf{Weak-supervised distance: }} 
    We propose the use of a labeling system to decide whether two patients should be grouped together, separated, or left undetermined, 
    and the use of a classifier, trained on such labels, to learn a distance.
    
\begin{figure}[ht]
        \centering
        \includegraphics[height=0.3\textheight]{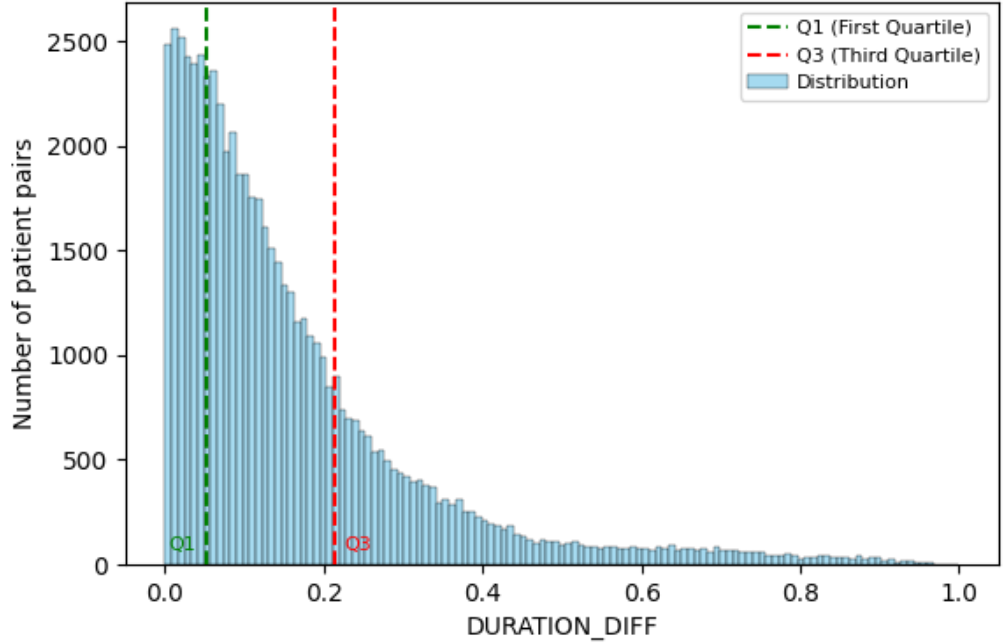}
        \caption{Distribution of the \texttt{DURATION\_DIFF} variable.}
        \label{fig:quartile_rule_example}
    \end{figure}

    To label patient pairs, we propose a voting-based system that relies on weak-supervised labelling functions 
    to assign labels. The design of the labelling functions was set up by frequent discussions with physicians.
    We devised seven functions, each based on a descriptive variable and relying on its distribution of values. A function assigns label \textit{grouped-together} (T) when the variable value is lower than first quartile, label \textit{separated} (S) when it is greater than third quartile, and label \textit{undetermined} (U) else (in this case the labelling function abstains, representing physicians uncertainty). This is illustrated in Fig.~\ref{fig:quartile_rule_example}.
    
\begin{figure}[ht]
        \includegraphics[height=0.185\textheight]{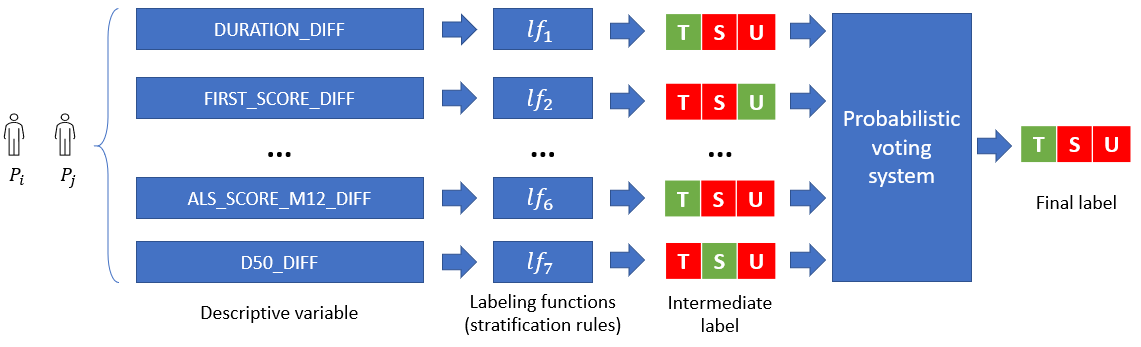}
        \caption{Example of pair labeling: Each function labels a pair, and the final decision is made by a probabilistic voting system. T, S, and U indicate whether the pair should be grouped, separated, or remain undetermined. The assigned label is shown in green.}
        \label{fig:vote_functions_details}
    \end{figure}

    The final label is determined by a probabilistic voting system across all functions (Fig.~\ref{fig:vote_functions_details}) \cite{ratner2017snorkel}. The system models the accuracy and correlations of each labeling function, learning their reliability without ground-truth labels. It then aggregates the outputs into a probability distribution, selecting the label with the highest probability as the final assignment.
    
    To learn our measure, we use a linear Support Vector Machine (SVM) classifier. It is trained on labeled pairs provided by the voting-based system in order to learn the weights assigned to each proposed labeling function for each descriptive variable. We define our measure named Weak-Supervised Distance (WSD)
    as follows:
    $$
    \mathit{WSD}(X, Y) = \sum_{i=1}^k w_i \cdot |x_i-y_i|
    $$
    where: $w_{i}$ denotes the weight associated with the $i^{th}$ labeling function,
    and $|x_i-y_i|$ is the value of the descriptive variable.

    The absolute value of \( w_{i} \) reflects the discriminative power of the function 
    (a high value indicates that it 
    is very influential in the decision-making process), while the sign of \( w_{i} \) indicates whether the function
    favors the grouping of patients or, conversely, their separation. Pairs labeled as U by the voting-based system are not considered in the learning of the weights.
    \section{Experiments}
    \label{sec:experiments}
    
This section describes our experiments for evaluating the proposed distances, comparing them to state-of-the-art approaches, and analysing cluster quality.

\paragraph{Setting.} 
    The experiments are conducted using the data analysis software Konstanz Information Miner (KNIME, version 5.4.2) in combination with the Python programming language (version 3.11.11). The probabilistic voting system is implemented using the generative model Snorkel (version 0.10.0) \cite{ratner2017snorkel}, which provides dedicated Python libraries. The labeling functions are thus defined within this model. To learn parameters of the sigmoid, we use the curve\_fit library that use non-linear least squares to fit the ALSFRS-R sequence \cite{virtanen2020scipy}.
    
\paragraph{Dataset.}    
    We work with data from 888 \guillaume{ALS} 
    patients diagnosed with motor neuron disorders, who are followed typically every 3 months at the university regional hospital center of Tours. \guillaume{No medical criteria were used to filter patients.} The progression is measured using the ALSFRS-R score. We exclude:
        
    \begin{enumerate}
        \item Patients with fewer than five visits (i.e., $n < 5$)
        to ensure regular follow-up. This leads to the exclusion of 350 patients.
        \item Patients with sequences considered abnormal by the physicians. This includes sequences where the ALSFRS-R score increases by more than two points between two consecutive visits (i.e., $s_{t_{i+1}} - s_{t_{i}} > 2$), a clinically extremely unlikely scenario, as patients' health tends to deteriorate over time. Therefore, 151 patients are excluded. Additionally, after presenting a detailed visualization of the ALSFRS-R sequences to the physicians, three more patients were excluded due to their sequence being deemed impossible by the experts.
        \item Patients whose first recorded ALSFRS-R score arrives 
        more than 30 days after their first visit are excluded,
        avoiding any bias related to duration. This leads to the exclusion of 19 patients.
        \item Finally, since we also compare our approach to the HAL approach, which uses ALSFRS subscores, patients without these available subscores are excluded. This filter excludes 12 patients.
    \end{enumerate}

    At the end, 353 patients are retained, leading to 62,481 pairs. Spearman correlation test did not reveal any highly correlated descriptive variables (with a correlation coefficient $\geq 0.7$). Therefore, we retain all variables described in Section \ref{sec:extracted_features} to develop the distances. 
    
\paragraph{Sigmoid.}
    Using the best-performing sigmoid function, the model achieves a mean error below 2 on the ALSFRS-R score for the majority of patients in the cohort. An error margin comparable to the variability occasionally observed among physicians. The median error is 1.43, indicating that for half of the population, the prediction error remains below this value. This corresponds to a deviation of approximately $\frac{1.43}{48} \approx 3\%$, which is minimal and therefore deemed acceptable for our experiment.
    
\paragraph{Distance properties check.}
    
    \guillaume{The properties of the WSD and cosine distance measures have been empirically evaluated over the dataset. For both measures, positivity, symmetry, and identity (i.e., the distance from a patient to itself is zero) are satisfied, as expected. Regarding  triangle inequality, empirical tests show that violations are rare and very minor in magnitude. Specifically, WSD satisfies the triangle inequality in 98.309\% of cases and cosine distance in 71.37\%, with WSD showing significantly smaller deviations (i.e., between 1E-06 and 1E-02) when violations occur. It is also worth noting that clustering using non-distances measures is a common and accepted practice in the literature.} 
    
\paragraph{Protocol.}
    We compare our method (i.e., clustering and distance), to state-of-the-art methods GOM, GRO, MEY and HAL, described in Section \ref{sec:sota}.
    
    We precompute and store measures in a matrix, which serves as input for various \guillaume{off-the-shelf} clustering methods. To enhance cluster detection, we also explore the use of a dimensionality reduction technique on the matrix. Concretely, we use Uniform Manifold Approximation and Projection (UMAP). UMAP achie-ves a compromise between preserving local and global structures in the projected data, and the appropriate choice of its parameters can induce space distortions that are beneficial for 
    \guillaume{cluster} identification \cite{mcinnes2018umap}. 
    \guillaume{The} clustering methods are then applied to the resulting low-dimensional representations. To evaluate the impact of dimensionality reduction, we compare clustering performance with and without UMAP. Furthermore, we investigate the impact of using different \guillaume{off-the-shelf} clustering methods. The following clustering methods are considered:

    \begin{itemize} 
        \item \textbf{K-means (KME)}: Applied only to UMAP representations, as it requires vector space coordinates to compute centroids. 
        \item \textbf{K-medoids (KMD)}: Applied to both precomputed matrices and UMAP representations. 
        \item \textbf{Agglomerative Hierarchical Clustering (AHC)}: Applied to both precomputed matrices and UMAP representations, using the "complete" linkage criterion to favor the formation of compact clusters.
    \end{itemize}
    
    \guillaume{Each of theses possibles combinations distance + 
    clustering method, are called workflows. }
    For each \guillaume{workflow},
    we vary the number of clusters, testing from 2 to 6. This yields to 100 
    \guillaume{workflows} to assess.
    
    For the state-of-the-art approaches, we use the setting recommended by authors. Approaches using thresholds-based rules, GOM, GRO and MEY, are not paired with 
    \guillaume{all}
    clustering methods nor UMAP and have fixed number of clusters. HAL approach uses AHC, varying between 4 and 6 clusters.
    Additionally, we investigate whether applying UMAP to their dissimilarity matrix may produce better results. This yields to 9 additional 
    \guillaume{workflows}
    to assess.
    
\paragraph{Cluster quality.}    
    To evaluate the quality of the resulting clusters from different 
    \guillaume{workflows}, we propose 3 criteria:
    
    \begin{enumerate}
        
        \item \textbf{Survival analysis:} We assess whether the identified clusters exhibit distinct survival curves.
        To compare the survival curves of different clusters, we apply the log-rank test to each pair of curves \cite{kleinbaum2012kaplan}. The log-rank test evaluates whether the survival curves of two 
        \guillaume{clusters} are significantly different. The null hypothesis of this test assumes that there is no difference in survival between the two \guillaume{clusters}, i.e., likelihood of survival at any given time is the same for both \guillaume{clusters}. A p-value below 0.05 indicates a significant survival difference, suggesting that the two survival curves differ.
        
        The test also provides a Log-Rank Statistic (LRS), calculated as:

        $$
        LRS = \frac{(O - E)^2}{Var(O-E)}
        $$
        
        where $O$ and $E$ represent the observed and expected number of events (i.e. death of patient) at a particular time point, and $Var$ is the variance that reflects how much variability we expect in the number of events across time. 

        We apply this test to every pair of curves and report the maximum p-value and minimum LRS among all comparisons.
        P-values range from 0 to 1, where smaller values suggest greater significance.
        LRS range from 0 to +$\infty$, with higher values indicating better separation between the survival curves, reinforcing the clinical relevance of the findings.
        
        \item \textbf{Silhouette score:} Silhouette score is an internal metric that favors compact, well-separated clusters~\cite{rousseeuw1987silhouettes}. For each patient, it measures the difference between cohesion (average distance to patients in the same cluster) and separation (average distance to the nearest cluster)~\cite{shahapure2020cluster}. The silhouette score ranges from -1 to +1: a positive score indicates good clustering, with values close to +1 suggesting well-separated and well-defined clusters; a score near 0 indicates overlapping clusters, and a negative score indicates misclassification.
        
        \item \textbf{Patient distribution:} We also suggest analyzing how patients are distributed among the clusters by counting the number of patients per cluster, evidencing imbalanced clusters. This is important, as highly uneven clusters with very few patients may indicate outliers and make it challenging to draw meaningful conclusions.
    \end{enumerate}
    
\paragraph{Results.}   
    We executed the 109 mentioned 
    \guillaume{workflows}
    on the patients dataset and assessed the described quality criteria.
    To reduce the analysis of results, we focus on 
    \guillaume{workflows}
    with a silhouette score of 0.5 or higher. This keeps 50 
    \guillaume{workflows}, 41 concerning the proposed measures and 9 from state-of-the-art methods. To align with the log-rank test, we further filter out 
    \guillaume{workflows}
    with non-significant p-values (p-value $\geq$ 0.05), keeping 
    \guillaume{24}
    \guillaume{workflows}, 
    \guillaume{21} from our approach, and 3 from state-of-the-art approaches. We remark that all \guillaume{workflows} 
    using the HAL approach are not significant.
    
    We present the results in Table \ref{tab:combination_results}, which includes the number of clusters, the number of patients in each cluster, the silhouette score, and the log-rank test with its p-value and LRS. The results are sorted in descending order of the LRS.
    \guillaume{Workflow} names follow the format: `measure'\_`UMAP, if applied'\_`
    clustering method'\_`number of clusters'. State-of-the-art methods are represented in bold.
    
    We observe that 
    \guillaume{workflows}
    producing five or six clusters are excluded
    due to insufficient statistical significance in survival distribution. 
    In addition, most of the reported 
    \guillaume{workflows}
    primarily involve UMAP, as those 
    without UMAP result in silhouette scores below 0.5. 
    \guillaume{Workflows} concerning
    the MAN measure are also filtered out.

    The best-performing 
    \guillaume{workflow}
    is COS\_KMD\_2, which achieved 
    the most statistically significant log-rank p-value (6.92E-29), along with a LRS of 124.39. It also achieves the highest silhouette score (0.79) but its high standard deviation (± 0.39) indicates instability in cluster cohesion, suggesting that conclusions on silhouette
    should be drawn with caution.
    
    Other notable 
    \guillaume{workflows}
    include WSD\_UMAP\_AHC\_2 and GOM\_2, which yield high and comparable LRS (110.69 and 106.32, respectively) and relatively robust silhouette scores ($\geq$ 0.57). Overall, \guillaume{workflows}
    involving WSD with UMAP followed by AHC, KME, or KMD offer a favorable compromise between cluster compactness and clinical separation, particularly in two and three clusters settings.
    
    Interestingly, COS\_AHC\_2 attains the highest silhouette score overall (0.93 ± 0.17), yet exhibits a substantially lower LRS (33.09), suggesting that while the clusters are well separated in the descriptive variables space, they lack meaningful prognostic differentiation. Clinical feedback from collaborating physicians further confirms that the patient groupings produced by this 
    \guillaume{workflow}
    is not clinically relevant or interpretable \guillaume{
    \guillaume{as there are}
    very few patients in \vero{the} second cluster}. 
    Additionally, configurations involving more than two clusters generally result in diminished LRS, underscoring a decline in clinical 
    \guillaume{clustering} power with increased cluster granularity.
    
    In summary, WSD\_UMAP\_AHC\_2, GOM\_2 and COS\_KMD\_2, achieve the most effective balance between intra-cluster consistency and clinical relevance with a high silhouette average. However, the silhouette score for COS-\_KMD\_2 exhibits a high standard deviation, so we should interpret the results with caution and refrain from drawing definitive conclusions for this \guillaume{workflow}. While increasing the number of clusters often dilutes statistical power, most of proposed 
    \guillaume{workflows}
    for three, and four clusters consistently outperform state-of-the-art techniques (i.e., MEY\_3 and GRO\_4) in terms of LRS, without compromising silhouette quality. 
    
    In addition to the proposed metrics, physicians' \guillaume{feedback} confirms that for two, three, and four clusters, the first two 
    \guillaume{workflows}
    (i.e., COS\_KMD\_2, WSD-\_UMAP\_AHC\_2, EUC\_UMAP\_KMD\_3, EUC\_UMAP\_KME\_3, COS\_U-MAP\_KME\_4 and COS\_UMAP\_KMD\_4)  produce clinically coherent clusters that are generally well-sized and show distinct progression patterns in the survival curves. \guillaume{These findings are promising for the clinical care and the clustering of future patients, as they provide clinically meaningful, well-sized clusters with clearly distinct survival curves.}
    
    
    These results must be interpreted with caution. In particular, although silhouette metric offers useful insights into cluster cohesion, it is inherently influenced by the distribution of distances specific to each metric. As such, comparing silhouette values across different distance measures may confound clustering quality with properties of the distance space itself. Therefore, we place greater emphasis on survival-based metrics like log-rank test, that more directly reflect clinical \guillaume{clustering}\vero{.}

\newcolumntype{P}[1]{>{\centering\arraybackslash}p{#1}}

\begin{table}[t]
\centering
\resizebox{\textwidth}{!}{%
\begin{tabular}{|c|P{1.5cm}|P{2.5cm}|c|P{1.5cm}|P{1.5cm}|}
\hline
\textbf{
\guillaume{Workflow} name} & \textbf{Number of clusters} & \textbf{Number of patients per cluster} & \textbf{Silhouette} & \textbf{Log-rank p-value (max)} & \textbf{LRS (min)} \\
\hline
COS\_KMD\_2 & 2 & 298, 55 & 0.79 $\pm$ 0.39 & 6.92E-29 & 124.39 \\
WSD\_UMAP\_AHC\_2 & 2 & 236, 117 & 0.57 $\pm$ 0.23 & 6.91E-26 & 110.69 \\
\textbf{GOM\_2} & 2 & 245, 108 & 0.61 $\pm$ 0.23 & 2.92E-21 & 106.32 \\
WSD\_UMAP\_KME\_2 & 2 & 181, 172 & 0.56 $\pm$ 0.20 & 2.22E-20 & 85.58 \\
WSD\_UMAP\_KMD\_2 & 2 & 186, 167 & 0.56 $\pm$ 0.20 & 5.86E-18 & 74.57 \\
WSD\_AHC\_2 & 2 & 328, 25 & 0.56 $\pm$ 0.23 & 2.95E-17 & 71.38 \\
EUC\_UMAP\_KME\_2 & 2 & 183, 170 & 0.50 $\pm$ 0.21 & 2.59E-15 & 62.56 \\
EUC\_UMAP\_AHC\_2 & 2 & 225, 128 & 0.50 $\pm$ 0.23 & 6.18E-15 & 60.84 \\
EUC\_UMAP\_KMD\_2 & 2 & 187, 166 & 0.50 $\pm$ 0.21 & 4.84E-14 & 56.79 \\
COS\_AHC\_2 & 2 & 348, 5 & 0.93 $\pm$ 0.17 & 8.80E-09 & 33.09 \\
EUC\_UMAP\_KMD\_3 & 3 & 131, 121, 101 & 0.50 $\pm$ 0.19 & 4.66E-08 & 29.85 \\
EUC\_UMAP\_KME\_3 & 3 & 132, 123, 98 & 0.50 $\pm$ 0.19 & 1.10E-07 & 28.19 \\
WSD\_UMAP\_AHC\_3 & 3 & 143, 117, 93 & 0.61 $\pm$ 0.15 & 3.84E-06 & 21.35 \\
WSD\_UMAP\_KME\_3 & 3 & 142, 118, 93 & 0.61 $\pm$ 0.15 & 4.79E-06 & 20.92 \\
WSD\_UMAP\_KMD\_3 & 3 & 141, 119, 93 & 0.61 $\pm$ 0.15 & 5.92E-06 & 20.51 \\
COS\_UMAP\_KME\_4 & 4 & 133, 80, 74, 66 & 0.59 $\pm$ 0.17 & 3.90E-05 & 16.92 \\
COS\_UMAP\_KMD\_4 & 4 & 133, 78, 74, 68 & 0.59 $\pm$ 0.17 & 3.90E-05 & 16.92 \\
\textbf{MEY\_3} & 3 & 230, 97, 26 & 0.59 $\pm$ 0.21 & 5.13E-05 & 16.40 \\
COS\_UMAP\_AHC\_4 & 4 & 141, 85, 66, 61 & 0.57 $\pm$ 0.22 & 3.08E-04 & 13.02 \\
WSD\_UMAP\_AHC\_4 & 4 & 143, 93, 92, 25 & 0.55 $\pm$ 0.24 & 6.17E-04 & 11.72 \\
EUC\_UMAP\_KMD\_4 & 4 & 106, 97, 78, 72 & 0.54 $\pm$ 0.18 & 5.32E-03 & 7.77 \\
COS\_KMD\_3 & 3 & 218, 116, 19 & 0.62 $\pm$ 0.40 & 6.10E-03 & 7.52 \\
\textbf{GRO\_4} & 4 & 143, 116, 82, 12 & 0.54 $\pm$ 0.22 & 2.01E-02 & 5.41 \\
COS\_KMD\_4 & 4 & 139, 129, 68, 17 & 0.52 $\pm$ 0.38 & 3.18E-02 & 4.61 \\
\hline
\end{tabular}
} 
\caption{Results of different 
\guillaume{workflows}
with silhouette and log-rank. State-of-the-art 
\guillaume{workflows}
are highlighted in bold.}
\label{tab:combination_results}
\end{table}
    \section{Conclusion}
    \label{sec:conclusion}
    We introduce a method to compare ALSFRS-R sequences using both simple distance metrics and an advanced weakly-supervised approach combining a probabilistic voting system with a linear SVM. The resulting distance matrices are paired with a\guillaume{n} \guillaume{off-the-shelf} clustering method.

    We propose different workflows by varying the number of clusters, 
    clustering methods, and incorporating UMAP for dimensionality reduction. In total, we evaluated 100 workflows and compared to 9 workflows from state-of-the-art methods. Our approach consistently achieves silhouette scores comparable to state-of-the-art techniques and yields higher log-rank statistics for two, three and four clusters.
    
    These experiments show that several workflows, achieve a strong balance between cluster cohesion and clinical relevance. Physicians' feedback further support that these workflows generate clinically meaningful, well-sized clusters with clearly distinct survival curves. This is a promising result for the clinical management and 
    \guillaume{the clustering} of future patients.

    
    
    \guillaume{Our study shows that high internal clustering metrics like the silhouette score don’t always reflect clinical relevance, as survival analysis reveals. While useful for assessing cluster structure, the silhouette score should be combined with clinical metrics to ensure meaningful evaluation.} 

    We identify \guillaume{five} key \guillaume{directions} 
    for future work. The first one is the integration of the temporal dimension in defining patient clusters. Currently, our method aggregates ALSFRS-R sequences by precomputing descriptive variables, which are then used in 
    clustering methods. However, given the heterogeneous nature of the disease, it is likely that a patient’s 
    \guillaume{cluster}
    changes over time. Thus, analyzing these variables across different time windows (e.g., from baseline to 3 months, from 3 to 6 months, etc.) could provide more relevant insights. Incorporating this temporal aspect would enhance the clinical relevance of the patient clusters.
    
    The second \guillaume{key direction}
    is improving the quality of the resulting clusters. Many 
    studies do not assess the quality of their clusters in-depth and often overlook internal metrics, such as the silhouette score, to evaluate cluster separation. Typically, 
    \guillaume{clusters} are compared by analyzing statistical differences in clinical variables \cite{tavazzi2023artificial} after 
    \guillaume{clustering}. To strengthen the reliability of 
    \guillaume{clustering} methods, it is essential to incorporate both internal and external evaluation metrics. A promising direction for external evaluation could be the use of biological data to identify parameters that influence the disease progression.
    
    \guillaume{The third key direction is about the representativeness of the dataset. It is important to ensure that the distribution of descriptive variables is properly verified before normalization. In particular, minimum and maximum values should be checked to detect potential outliers. Additionally, the required number of patients should be assessed to ensure the dataset accurately reflects the target population. Statistical methods, such as Cochran's formula \cite{cochran1977sampling}, may be considered to estimate an appropriate sample size. Failing to do so could distort the distribution, ultimately leading to a non-representative dataset.}
    
    \guillaume{The fourth key direction is to adapt our approach for use in other diseases characterized by progressive and heterogeneous progressions such as Parkinson’s, multiple sclerosis, or cancer. To support these diverse disease contexts, we aim to generalize our approach to any sequences with quasi-monotonic behavior, by accommodating various longitudinal clinical or biological parameters.}
    

    
    Finally, we plan to conduct experiments on larger datasets to validate and enhance the robustness of our findings. PRO-ACT appears to be the most suitable option for external validation, as it has been widely used in studies such as GRO, GOM, and HAL. Although it has known limitations, PRO-ACT is widely recognized and accessible, offering better consistency by including more patients than the current dataset. Using PRO-ACT will allow us to compare our results with previous studies, ensuring that our findings are generalizable and reliable despite the imperfections of the data.

\begin{credits}
    \subsubsection*{\discintname}
    The authors have no competing interests to declare that are relevant to the content of this article.
    \guillaume{\subsubsection*{Availability of Data and Materials.}
    Data and tools used for experiments are available via this link: \underline{\href{https://github.com/GuillaumeTejedor/WSD.git}{https://github.com/GuillaumeTejedor/WSD.git}}}.
\end{credits}
    
    \bibliographystyle{splncs04}


\end{document}